%% file: main.tex
\pdfoutput=1
\documentclass[11pt]{article}

\usepackage[final]{acl}

\usepackage{times}
\usepackage{latexsym}

\usepackage[T1]{fontenc}
\usepackage[utf8]{inputenc}

\usepackage{microtype}

\usepackage{inconsolata}

\usepackage{graphicx}
\usepackage{soul}

\usepackage{tikz}
\usetikzlibrary{shapes, arrows.meta, positioning}

\title{Leveraging Graph Structures to Detect Hallucinations in Large Language Models}

\author{
 \textbf{Noa Nonkes\thanks{Equal contribution.}},
 \textbf{Sergei Agaronian\footnotemark[1]},
 \textbf{Evangelos Kanoulas},
 \textbf{Roxana Petcu}
\\
\\
 University of Amsterdam
\\
 \small{
   noanonkes@gmail.com,
   r.m.petcu@uva.nl
 }}

\usepackage{booktabs}
\usepackage{multirow}
\usepackage{amsfonts}
\usepackage{amsmath}

\begin{document}
\maketitle

\begin{abstract}
  \input{content/abstract}
\end{abstract}

\section{Introduction}
\input{content/introduction}

\section{Related Work}\label{sec:related_work}
\input{content/related_work}

\section{Methodology}
\input{content/method}

\section{Experimental Setup}
\input{content/empirical}

\section{Results}
\input{content/results}

\section{Conclusion}
\input{content/conclusion}

% \bibliography{custom}
\bibliography{main}

\appendix

\section{Appendix}
\label{sec:appendix}

% Figure environment to prevent page break
\begin{figure}[htbp]
\centering
\caption{Prompt 1: Without provided Context}
\begin{minipage}{\linewidth} % Minipage to keep the content together
% Dotted line
\noindent\makebox[\linewidth]{\dotfill}
\textbf{Task:} \{\textit{Imagine you are crafting a multiple-choice exam in the field of biomedical studies.}\} \\

\{\textit{Your task is to generate a set of statements related to a given question.}\} \\

\{\textit{Provide one accurate statement as the correct answer (Answer 1) and four misleading statements that should appear as plausible distractors (Answers 2 to 5).}\} \\

\{\textit{Ensure that the incorrect answers are not easily mistaken for accurate information related to the question.}\} \\

\textbf{Question:} \{\textit{What is the role of the BRCA1 gene in breast cancer?}\} \\

\textbf{Context:} \{\textit{}\} \\
\noindent\makebox[\linewidth]{\dotfill}
\end{minipage}
\end{figure}

% Figure environment to prevent page break
\begin{figure}[htbp]
\centering
\caption{Prompt 2: With provided Context}
\begin{minipage}{\linewidth} % Minipage to keep the content together
% Dotted line
\noindent\makebox[\linewidth]{\dotfill}
\textbf{Task:} \{\textit{Imagine you are crafting a multiple-choice exam in the field of biomedical studies.}\} \\

\{\textit{Your task is to generate a set of statements related to a given question.}\} \\

\{\textit{Provide one accurate statement as the correct answer (Answer 1) and four misleading statements that should appear as plausible distractors (Answers 2 to 5).}\} \\

\{\textit{Ensure that the incorrect answers are not easily mistaken for accurate information related to the question.}\} \\

\textbf{Question:} \{\textit{What is the role of the BRCA1 gene in breast cancer?}\} \\

\textbf{Context:} \{\textit{The BRCA1 gene is a gene on chromosome 17 that produces a protein responsible for repairing DNA. Mutations in this gene can lead to reduced protein functionality, impairing DNA repair processes. This impairment increases the risk of mutations in other genes, which can result in uncontrolled cell growth and potentially lead to the development of breast cancer. The presence of mutated BRCA1 is a significant marker for an increased risk of breast and ovarian cancers in women, making genetic testing a key preventive measure for those with a family history of these cancers.}\} \\

\noindent\makebox[\linewidth]{\dotfill}
\end{minipage}
\end{figure}

\end{document}

%% file: content/abstract.tex
% Large language models are extensively applied across diverse areas of expertise, such as household management, financial consulting and healthcare guidance. However, a well-known drawback of LLM-driven systems is their predisposition to generate hallucinations. We propose a method to detect misinformation in retrieval-driven LLM generations using graph structures and semantic similarity. We leverage Graph Attention Networks (GATs) to compute dependencies between true statements and misinformation. Our findings show that GATs outperform conventional neural approaches, establishing a framework for learning semantically rich sentence dependencies and offering a promising direction for misinformation detection. 
% The experiments further indicate that leveraging contrastive learning significantly enhances performance. When evaluated against open-book benchmarks, our model comes close in performance without access to search-based methods.\footnote{The full code can be found on our \href{https://anonymous.4open.science/r/Misinformation-Detection-in-LLMs-ANON/}{GitHub repository}.}

Large language models are extensively applied across a wide range of tasks, such as customer support, content creation, educational tutoring, and providing financial guidance. However, a well-known drawback is their predisposition to generate hallucinations. This damages the trustworthiness of the information these models provide, impacting decision-making and user confidence. We propose a method to detect hallucinations by looking at the structure of the latent space and finding associations within hallucinated and non-hallucinated generations. We create a graph structure that connects generations that lie closely in the embedding space. Moreover, we employ a Graph Attention Network which utilizes message passing to aggregate information from neighboring nodes and assigns varying degrees of importance to each neighbor based on their relevance. Our findings show that 1) there exists a structure in the latent space that differentiates between hallucinated and non-hallucinated generations, 2) Graph Attention Networks can learn this structure and generalize it to unseen generations, and 3) the robustness of our method is enhanced when incorporating contrastive learning. When evaluated against evidence-based benchmarks, our model performs similarly without access to search-based methods.\footnote{The full code can be found on our \href{https://github.com/noanonkes/Hallucination-Detection-in-LLMs}{GitHub repository}.}

% ROXANA:
% The Abstract is usually difficult to write. I suggest you look at some examples of papers on LLM hallucinations, and try to mimic their structure (for example: one sentence on LLMs, one sentence on hallucinations being a problem, *two sentences about what you do, two sentences on main results, and one ending sentence).

% *one sentence about what you do should start with "In this paper, we propose..." or simply "We propose", "We look into", ...

% Here are some examples of good abstracts:

% https://dl.acm.org/doi/10.1145/3583780.3614905
% https://arxiv.org/pdf/2310.01469.pdf
% https://arxiv.org/pdf/2303.08896.pdf

%% file: content/introduction.tex
%% Introduction
Large Language Models (LLMs) have recently surged in popularity, notably due to the emergence of agents and models such as ChatGPT, Bard, Vicuna, and LLaMA~\cite{radford2019language, pichai2023important, vicuna2023, LLAMA2}. 
Despite their increased capabilities for complex reasoning~\cite{radford2019language}, substantial challenges persist in grounding LLM generations to verified real-world knowledge. Ensuring that LLM generations are not only plausible but also factually correct poses a complex problem~\cite{xu2024hallucinationinev}, which can be minimized~\cite{liang2024} but so far not eliminated. Needless to say, even though LLMs possess an unparalleled ability to produce fast, credible, and human-like output, they are prone to hallucination \cite{Ji_2023, KASNECI2023102274}. This challenge expresses the non-trivial need for robust methods that detect and mitigate the spread of LLM-generated hallucinations.

%% Motivation
\paragraph{Motivation} The first underlying premise of this study is that LLM hallucinations are not unstructured, i.e., hallucinations share characteristics in the latent space. While extensive work has been done to mitigate LLM hallucinations~\cite{ji2023sol, feldman2023sol, martino2023sol}, identifying these by their structural properties remains largely unexplored. ~\citet{selfcheckGPT} make a first step into identifying model-agnostic hallucinations through their SelfCheckGPT method, reliant solely on black-box access to the model to infer new generations. % while other methods require access to either the the LLM's logits, or an external knowledge repository. 
This method aligns with the idea that, given the same query, non-hallucinated samples exhibit a higher degree of similarity with each other than with hallucinated samples. We aim to extend this exploration by analyzing if, \textbf{independent} of the query, hallucinations share a higher degree of similarity with each other than with non-hallucinated generations. While SelfCheckGPT employs an implicit approach to model consistencies, we aim to do it explicitly by exploring semantic correspondences between hallucinations in the latent space. The second premise relies on the principle of homophily, which expresses that entities that share similar characteristics are more likely to form connections with each other ~\cite{zhang2016homophily}. In the context of hallucinations and text representations, homophily suggests that samples that share text-level characteristics tend to lie closer in the embedding space. We study if the degree of hallucination is such a characteristic.

%% Why graphs
Based on the outlined premises and assumptions, we propose leveraging graph structures and message passing to reveal underlying patterns in the data. Comparing sentences in the embedding space involves assessing pairwise similarities between any pair of sentences. This results in $N(N-1)/2$ computations for $N$ sentences, which is not computationally expensive for a one-time calculation, however, applying a neural network on top of this structure does not scale computationally, even with a simple single-layer feed-forward network. However, by leveraging the principle of homophily, we can form a graph where only similar nodes will have a direct connection between them, significantly reducing computational costs.

%% Hypothesis and research questions
\paragraph{Objectives} This study proposes two hypotheses: 1) LLM hallucinations arise from a pattern, which reflects in shared characteristics in the embedding space, and 2) we can efficiently leverage these characteristics using graph structures. We can formulate the following research questions:

\begin{enumerate}
    \item Do LLM-generated hallucinations share characteristics?
    \item Can we leverage graph structures to identify and learn these characteristics?
    \item If learned, can we use this knowledge to identify hallucinations among new incoming LLM generations through label recovery?
\end{enumerate}

%% We introduce...
\paragraph{Contributions} We introduce a hallucination detection framework for LLM-generated content. Given an existing dataset of hallucinations and true statements, we 1) leverage semantically rich sentence embeddings, 2) construct a graph structure where semantically similar sentences are connected, 3) train a Graph Attention Network (GAT) model that facilitates message passing, neighborhood attention attribution and selection, and 4) employ the GAT model to categorize new sentences as hallucinated or non-hallucinated statements.

%% We find that...
According to our findings, 1) using semantic information to form connections between entities in the latent space helps to uncover links within hallucinated and non-hallucinated statements, 2) non-local aggregation enhances these links, 3) contrastive learning helps in distinguishing embeddings and leads to better performance, and 4) our method can accurately classify new unseen sentences as hallucinations or true statements.

%% Goal not SOTA, datasets, and generalizability
The focus of this study is not getting on par performance with SOTA. Instead, we bring new hypotheses on the characteristics of LLM hallucinations. We implement and experiment with our method on multiple datasets: 1) we generate our own dataset by prompting an LLM to generate both true and misleading statements given a query and a context, and 2) we apply our framework to existing benchmark datasets to evaluate its performance on non-controlled data. Comparisons with benchmarks such as~\cite{selfcheckGPT, Thorne18Fever} show that our method achieves close performance. Notably, we do not need access to external knowledge, LLM logits, or additional inference passes to the LLM, while keeping computational costs minimal.

We hypothesize that the latent space holds rich information beyond features such as contextualized, syntactic, and semantic information, for which the embeddings have been previously trained to capture. We also hypothesize that this information can be discovered and leveraged using geometric information. We propose a method that can be extended beyond the hallucination problem, and which can be generalized, applied to, and experimented with using any categorical label.

\vspace{-1em}

%% file: content/related_work.tex
%%% Intro
\paragraph{LLMs} The field of Natural Language Processing (NLP) has seen a significant evolution, from early probabilistic approaches such as Naive Bayes~\cite{kim2006NaiveBayesApp} to transformer models~\cite{wolf2020Transformers} with attention mechanisms \cite{vaswani2017attention}. This evolution also leads to LLMs which play a significant role in NLP tasks and applications~\cite{GPT2, GPT3}, yet they face a significant challenge known as \textit{hallucination generation} that has been thoroughly studied.

% SelfCheckGPT
\paragraph{Prompt verification} SelfCheckGPT~\cite{selfcheckGPT} mitigates hallucinations using a sampling-based approach that facilitates fact-checking in a zero-resource fashion. The authors leverage the idea that if an LLM has knowledge on a certain subject, true generations from the same query are likely to be similar and factually consistent. They compare multiple query-dependent generated responses to identify fact inconsistencies indicative of hallucinations. Instead of only retrieving the most likely generated sequence of the model, they draw $N$ further stochastic LLM responses and query the model itself to ascertain whether each sample supports the hallucination. In essence, SelfCheckGPT does not need external knowledge but utilizes its internal knowledge to self-detect structural aspects of hallucinations. This approach, called prompt verification, achieves a notable 67\% AUC-PR score in factual knowledge classification, showing potential for zero-shot fact-checking. However, it involves computational overhead as sampling LLM generations requires multiple forward passes to classify each single statement. Outside of SelfCheckGPT, multiple other approaches leverage prompt verification \cite{dhuliawala2023COVe, varshney2023Stitch}. 

%%% FEVER
\paragraph{Retrieval-based} The Fact Extraction and Verification (FEVER) Shared Task~\cite{Thorne18Fever} brings together several other approaches to hallucination detection. The task participants were challenged to classify whether human-written facts can be supported or refuted while having access to documents retrieved from Wikipedia. The task is mostly split into three parts. For document selection, many teams adopt a multi-step approach, which typically involves techniques such as Named Entity Recognition~\cite{shaalan2014NER}, Noun Phrases~\cite{zhang2007NounPhrases}, and Capitalized Expression Identification. The results are then used as inputs for querying a search API such as Wikipedia. The next step involves extracting relevant sentences through methods such as keyword matching, supervised classification, and sentence similarity scoring. Finally, for natural language inference, the extracted evidence sentences are often concatenated with the claim and passed through models such as a simple multilayer perceptron (MLP), Enhanced LSTMs~\cite{chen2017EnhancedLSTM}, or encoder models to synthesize and evaluate the relationships between them. One notable difference between FEVER models and previously described work~\cite{selfcheckGPT} is that they have access to external sources. 

%%% TruthfulQA and why we don't use it
\paragraph{Benchmark datasets} TruthfulQA~\cite{truth} proposes a benchmark for analyzing how accurate a language model is in generating answers given a question. The benchmark includes 817 questions spanning over 38 categories, which require a wide range of reasoning capabilities, such as questions in health, law, finance, and politics. Moreover, the questions are crafted in a manner that could lead humans to provide incorrect answers due to false beliefs or misconceptions. In this work, the authors analyze the performance of models such as GPT-3, GPT-Neo, GPT-J, GPT-2 and T-5~\cite{GPT3, GPT-Neo, GPT2, t5transformer}, identifying that the best model was truthful on only 56\% of the generations, while human performance reaches 94\%. Compared with the other hallucination benchmarks, the correctness of an answer in TruthfulQA can only be assessed in association with its query. Therefore, we do not consider these answers as hallucinations on their own. While we could model this dataset by merging all answers with their associated queries, that would induce a major bias in the semantic similarity calculations when forming our method's graph structure. As a result, we do not evaluate our method on TruthfulQA but focus on datasets where hallucinations can be detected on the answer level only.
%%% If we wanna use TruthfulQA, we should: merge query+answer, get embeddings, model in graph, make prediction, calculate accuracy on the query level (accuracy of all 5 query+answer for a specific query; average across all queries) and then compare this with TruthX on MC1

%%% Others
\paragraph{Other hallucination detection methods} There are numerous alternative approaches. \citet{luo2023SelfFamiliarity} tests the familiarity of the LLM with the query prior to generation. The model withholds generation if the familiarity is low. Other studies look into Bayesian estimation in retrieval-augmented generation. \citet{wang2023Bayesian} achieves an AUC-PR of around 62\% for factual knowledge, but introduces additional time and compute due to reliance on a search engine for external evidence retrieval. Another approach~\cite{chen2023hallucination} aims to detect hallucinations through training a discriminator on the RelQA LLM-generated question-answering dialogue dataset. Their method achieves 85.5\% accuracy on automatic labels and 82.6\% AUC-PR on human labels, although reliance on human annotations introduces ambiguity.

%%% How is ours different
In comparison to previous work, our method does not require access to external knowledge, nor to the LLM used for generating data, avoids biases associated with additional prompting, and eliminates costs associated with further inference.

%% file: content/method.tex
Assume a dataset $D = \lbrace (x_i, y_i) \rbrace ^n_{i=1}$ consisting of $n$ samples, where $x_i$ denotes a sentence and $y_i$ represents an ordinal categorical label indicating the degree of hallucination of $x_i$. 
\subsection{Graph Construction}
Consider a model $\phi(x) = e$, where $e \in \mathbb{R}^{768}$, which maps $x_i$ to its sentence-level embedding representation $e_i$. We construct a graph $\mathcal{G}=(\mathcal{V}, \mathcal{E})$ as follows:

\begin{itemize}
    \item $\mathcal{V}$ is the set of nodes. Each node $v \in \mathcal{V}$ corresponds to a single data point $x_i$ from dataset $D$. More precisely, the features of a node consist of the sentence-level embedding $\phi(x_i)$. We employ BERT~\cite{BERT} as the model $\phi$ to transform textual representations into embeddings.
    \item Two utterances $\{u, v\} \in \mathcal{V}$ are connected with an edge $(u, v) \in \mathcal{E}$ if and only if the semantic similarity between nodes $u$ and $v$ exceeds a threshold $\tau$. The semantic similarity between two utterances is calculated using cosine similarity.
\end{itemize}

The choice of $\tau$ must ensure a balance in graph connectivity. Ideally, the node degree distribution should be relatively uniform, with a limited number of both highly connected and disconnected nodes. Additionally, we aim to avoid spikes in node degrees, as they may indicate the formation of hubs. The value of $\tau$ is dependent on the $D$.

\subsection{Graph Attention Network}\label{sec:gat}
We employ a GAT model \cite{GAT} on our semantically-driven graph structure $\mathcal{G}$. Computing attention scores over sentence embeddings involves significant computational costs due to their high-dimensional representation. Consequently, we first reduce the dimensionality of node features by training a basic MLP. Then, we apply GAT on the reduced node features. The model will learn to map the sentence embeddings to a label indicating its degree of hallucination. We choose to model this problem with graph structures for two primary reasons: 1) to aggregate information via message passing, driven by our intuition that sentences that exhibit similar degrees of factuality share common structural components, and 2) to leverage edge weights, such that the level of similarity influences the information shared through message passing is expected to influence the message passing mechanism. We formalize the problem as an ordinal regression task as follows:

\begin{itemize}
    \item \textbf{Label Encoding}: If a data point $x_i$ has associated label $L$, then it is classified into all lower labels. Let $L$ be an ordinal label and $\text{encode}(L)$ be the corresponding encoding. Then, we can define $\text{encode}(L)$ as:

    \begin{equation}
        \text{encode}(L)_i = \begin{cases}
        1 & \text{if } i \leq L \\
        0 & \text{otherwise}
        \end{cases}
    \end{equation}

    where $\text{encode}(L)_i$ represents the $i$th element of the encoding vector. 
\end{itemize}

%%%% something about label recovery
By employing a graph-based model, we aim to validate our assumptions that hallucinations exhibit shared characteristics within the embedding space. We add connections between sentences that show a high degree of similarity and, during training, we exchange information between nodes and their local neighborhood.

\subsection{Label Recovery Task}\label{sec:task}
Assume a new evaluation dataset $D'$. We first append $D'$ to the existing graph $\mathcal{G}=(\mathcal{V}, \mathcal{E})$, and then perform one forward pass to the trained GAT using the new graph $\mathcal{G'} = (\mathcal{V} \cup \mathcal{V'}, \mathcal{E} \cup \mathcal{E'} \cup \mathcal{A})$ to solve a label recovery task. $\mathcal{A}$ represents the edges formed between the nodes $V$ and $V'$. We define the label recovery task as follows:

\begin{itemize}
    \item \textbf{Label recovery}: Consider a graph $\mathcal{G}=(\mathcal{V}, \mathcal{E})$, where $\mathcal{V}$ is the set of vertices and $\mathcal{E}$ the set of edges. The label recovery task involves inferring missing labels for a subset of nodes $\mathcal{U} \subseteq \mathcal{V}$ based on available information of the known labels of nodes $\mathcal{V} \setminus \mathcal{U}$.
\end{itemize}

% TEST Counter({0: 2406, 2: 304, 3: 296, 1: 295})
% VAL Counter({0: 2382, 1: 319, 3: 308, 2: 291})
% TRAIN Counter({0: 11212, 2: 1405, 3: 1396, 1: 1386})

\subsection{Data Generation}
\label{sec:generation}
We first apply our method to our own hallucination-generated dataset. Creating our own dataset allows for more control over the modeling choices and requirements of our methodology, such as degree of connectivity, homophily, and encoding techniques, ensuring a more targeted evaluation. However, we recognize the importance of generalizability, and therefore we also validate our approach on existing datasets to assess its performance across diverse contexts and benchmarks. 

\paragraph{Prompt} During the process of LLM-driven hallucination generation, we design a prompt that guides the model to create statements for a multiple-choice exam, as shown in Appendix \ref{sec:appendix}. In our prompt, we refer to \textit{hallucinations} as \textit{misleading statements}. This is a modeling choice we took because \textit{hallucinations} and \textit{misleading statements} are conceptually similar, however, by labeling them as \textit{misleading} we guide the model to generate statements intended to deceive the reader into believing they are true. This approach ensures the generation of hard in-context hallucinations instead of general hallucinations.

\paragraph{Retrieval-based generation} We construct our data using retrieval-augmented generation on a representative question-answer (QA) dataset. We first sample queries along with their corresponding answer and associated context. We then instruct the LLM to generate a multiple-choice exam, where queries act as exam questions. This technique orients the LLM to generate misleading statements alongside true statements for each prompting stage, as a form of conditional generation. The evaluation of the generated data is conducted solely through human assessment. Although the exam-instruction format facilitates the generation of misleading statements, we acknowledge that LLMs are susceptible to bias and hallucinations, which can be reflected in our generated dataset. Existing biases in the model might skew the types of hallucinations in an unintended way, however, we assume that the effects of LLM hallucinations when intentionally induced have minimal impact on our study. We use two prompting techniques for instructing the LLM to generate both true and misleading answers, given either 1) the query, or 2) the query with its associated context, both part of the QA dataset. The latter is aimed at obtaining context-aware true statements from the LLM.

We thus generate a dataset that contains, for each query, 11 statements: 1 true \textit{extracted} from the QA dataset, 1 true without context \textit{generated}, 1 true with context \textit{generated}, and 8 hallucinated \textit{generated} statements. The overview of this process is illustrated in Figure \ref{fig:schematic}. We provide the prompts in Appendix \ref{sec:appendix}. Each statement is assigned a label $y_i \in [0,1,2,3]$ representing hallucinated, true w/o context, true w/ context, and true statement. We solve a categorical regression task, and as such, the labels are ordinal one-hot encoded as presented in Section \ref{sec:gat}.

\begin{figure}[htbp]
    \centering
    \includegraphics[width=\columnwidth]{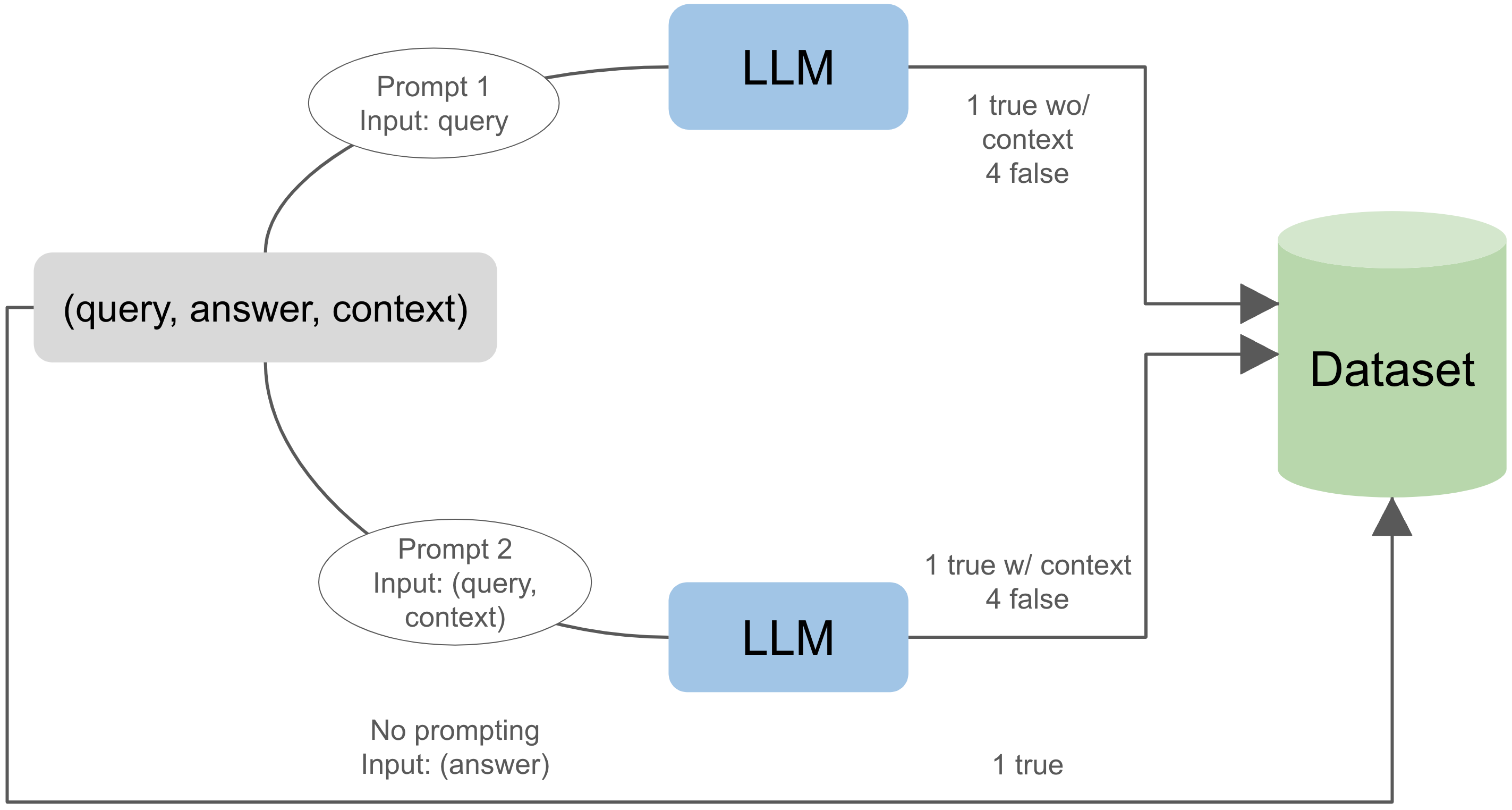}
    \caption{Data generation process.}
    \label{fig:schematic}
\end{figure}

% \begin{figure}[h!]
%     \centering
%     \includegraphics[width=\columnwidth]{images/prompting.png}
%     \caption{Data generation process.}
%     \label{fig:schematic}
% \end{figure}

%% file: content/empirical.tex
\subsection{Dataset} 
We used data from MSMARCO-QA \cite{msmarco}, selecting 2000 questions within the biomedical domain \cite{biomed} with answers consisting of a minimum of five words, to avoid sentences with explicit meaning only when paired with their corresponding question (examples are: "\textit{Yes}", "\textit{No}", "\textit{Non-surgical}", or "\textit{Virus infection}"). As mentioned in Section \ref{sec:generation}, each entry in the MSMARCO-QA dataset consists of the query, answer, and context. We use the queries and context to generate true w/ context, true w/o context, and hallucinated statements by prompting Meta's instruction-tuned 
Llama2 \cite{LLAMA2} $13B$ model\footnote{https://huggingface.co/OpenAssistant/llama2-13b-orca-8k-3319}. The dataset is randomly divided into three segments: training, validation, and test, with partitions of 70\%, 15\%, and 15\%, respectively. The random division is done on the sentence level, therefore same query generations are not necessarily in the same data partition.

\subsection{Graph}
We use the English uncased version of BERT \cite{BERT}\footnote{https://huggingface.co/bert-base-uncased} for sentence embeddings. The graph is constructed over the entire dataset, with designated masks for the training, validation, and test sets. Edges are formed between nodes with cosine similarity above a threshold $\tau = 0.85$, which was selected empirically to strike a balance, ensuring a reasonable level of graph connectivity. The resulting undirected graph has approximately $26M$ out of a potential of $240M$ edges (fully-connected). Additionally, we use the cosine similarity values as edge attributes.

% \begin{figure}[h!]
%     \centering
%     \includegraphics[width=1.1\columnwidth]{images/barplot_0.85.png}
%     \caption{Node connectivity.}
%     \label{fig:node_connectivity}
% \end{figure}

\subsection{Graph Attention Network} 
The embeddings are reduced to a dimensionality of 32 features using a trained single-layer MLP. The GAT model incorporates a single graph attention convolutional layer, transitioning from 32 to 3 dimensions, with 2 attention heads. The GAT model explores and aggregates information from the immediate neighborhood based on the attention mechanism. 

\subsection{Baselines} 
We employ a three-layer MLP with ReLU activations, using the same BERT 
model for sentence embeddings, but with concatenated query-answer as inputs. As a second baseline, we employ a larger pre-trained language model DeBERTa \cite{DEBERTA}\footnote{https://huggingface.co/MoritzLaurer/DeBERTa-v3-base-mnli-fever-anli}, processing query-answer pairs in a unified encoder for improved contextual understanding and predictions.
The MLP offers a simple yet powerful method for analyzing relationships between the generations and respective queries and has been shown to be efficient in sentence-level analysis~\cite{ramdhani2022sentiment, akhtar2017multilayer}. Meanwhile, DeBERTa is a complex transformer model that facilitates deeper comparisons of attention effects and structural differences between transformers and lightly connected graphs. Unlike our graph-based model, which relies solely on answer embeddings, the baselines utilize embeddings from both queries and answers, enhancing semantic expressiveness for better differentiation among classes.

\subsection{Training} The model selection process is based on optimal macro-recall performance on the validation split, offering a comprehensive evaluation of the model's ability to identify instances across all classes, crucial in the context of highly imbalanced data. Training spans over $500$ epochs utilizing the Adam optimizer~\cite{kingma2017adam} with fixed learning rate $1 \times 10^{-3}$. The model is trained to solve a categorical regression task using Binary Cross Entropy (BCE) loss~\cite{good1952rational}. This choice reflects the need for nuanced penalization of misclassifications, where the model applies a lower penalty for misclassifying adjacent, compared to further-apart labels. Efforts are made to prevent data leakage between the data partitions. During training, weights corresponding to edges that connect nodes with either validation or test nodes are nullified, ensuring no information exchange (Equation \ref{eq:train}). Backpropagation happens exclusively over the training nodes. Similarly, during validation, we modify edges connecting to test nodes (Equation \ref{eq:val}). Recall is calculated exclusively on validation nodes. The full graph is used to assess performance on the test set (Equation \ref{eq:test}).

{\small
\begin{align}
    \mathcal{N}(i) &= \{j|(i,j)\in\mathcal{E}, i\in\mathcal{N}_{train}\text{ and } \notag \\&\qquad j\in \mathcal{N}\backslash\{\mathcal{N}_{val}\cup\mathcal{N}_{test}\}\} \label{eq:train} \\
    \mathcal{N}(i) &= \{j|(i,j)\in\mathcal{E},i\in\mathcal{N}_{val}\text{ and } j\in \mathcal{N}\backslash\mathcal{N}_{test}\} \label{eq:val} \\
    \mathcal{N}(i) &= \{j|(i,j)\in\mathcal{E},i\in \mathcal{N}_{test}\} \label{eq:test}
\end{align}
}

\subsection{Evaluation Metrics}
We calculate three metrics to evaluate the performance of our approach. Macro-recall assesses the model's accuracy in identifying individual classes, while macro-precision evaluates prediction accuracy per class. AUC-PR calculates the area under the precision-recall curve, providing a measure for binary classification performance. Additionally, these metrics are robust against class imbalance, making them suitable for evaluating our model on our imbalanced dataset.

\subsection{Benchmark Datasets}\label{ssec:benchmark}
In our study, we utilize two human-generated and annotated benchmarking datasets, namely FEVER~\cite{Thorne18Fever} and SelfCheckGPT~\cite{selfcheckGPT}. Our method can be generalized across other domains. To apply our method to another dataset, we re-construct the graph with the respective data partitions and redefine the labeling accordingly.  These datasets are used for evaluating the performance of our models under conditions that mimic real-world scenarios. 
% More detailed descriptions of these datasets and 
The specific methodologies employed for their use are discussed in Section \ref{sec:related_work}.

The FEVER dataset \cite{Thorne18Fever} contains $185445$ claims which are divided into three types of claims, namely \textit{Supports}, \textit{Refutes}, and \textit{Not Enough Info}, each paired with evidence sentences. To apply our method to the FEVER dataset, we re-define the label as $y_i \in [0,1,2]$ and generate the graph based on the train/val/test partitions of FEVER. The participating models \cite{Thorne18Fever} formulate careful data processing approaches and make use of external sources to verify the factuality of the claims. If search-based evidence is found for a claim, it is classified as \textit{Supports} or \textit{Refutes}. Similarly, if no evidence is found, it is labeled as \textit{Not Enough Info}.

The SelfCheckGPT dataset \cite{selfcheckGPT} consists of 1908 sentences categorized as \textit{Accurate}, \textit{Minor inaccurate}, and \textit{Major inaccurate}. To apply our method to SelfCheckGPT, we again redefine the labels as $y_i \in [0,1,2]$ and generate the graph by randomly splitting the data into train/val/test sets.

%% file: content/results.tex
\subsection{Non-local Aggregation} 

To address our first research question, more specifically \textit{1. Do LLM-generated hallucinations share characteristics?}, we analyze if our framework identifies an underlying structure of the embedding space. As shown in Table \ref{tab:baselines}, GAT exhibits better performance compared to DeBERTa-QA and MLP-QA on all metrics. GAT has approximately 17\% higher recall than both baseline models on the validation set. This suggests its superior ability to identify positive instances, reduce false positives, and effectively differentiate between true and hallucinated statements.

\begin{table}[htbp]
\caption{Comparing performance between GAT, 3-layer MLP, and DeBERTa using query answer (QA). The best results for each metric and dataset split are highlighted in bold.}
\resizebox{\columnwidth}{!}{
\begin{tabular}{@{}lllll@{}}
\toprule
Split                  & Model                      & Recall & Precision & AUC-PR  \\ \midrule
\multirow{3}{*}{Train} & GAT                        & \textbf{0.5069} & \textbf{0.5844} & \textbf{0.4153}  \\ 
                       & DeBERTa-QA                      & 0.3882 & 0.5404 & 0.3517 \\ 
                       & \multicolumn{1}{l}{MLP-QA} & 0.3214 & 0.3880 & 0.2718 \\ \midrule
\multirow{3}{*}{Val}   & GAT                        & \textbf{0.4972} & \textbf{0.5717} & \textbf{0.4096} \\ 
                       & DeBERTa-QA                      & 0.3206 & 0.5059 & 0.3357  \\ 
                       & \multicolumn{1}{l}{MLP-QA} & 0.3150 & 0.3622 & 0.2953  \\ \bottomrule
\end{tabular}
}
\label{tab:baselines}
\end{table}

\subsection{Contrastive Learning}\label{sec:CL}
Initial experiments revealed that BERT embeddings are not discriminative enough for our task. This is intuitively to be expected: we hypothesize that hallucinations share features in the latent space. However, this does not imply that these features are inherently discriminative within BERT embeddings, as BERT is trained to capture contextual, syntactic, and semantic information, rather than “validity” or “truthfulness”. 

To acquire enriched embeddings, we train a Contrastive Learning (CL)~\cite{khosla2021supervised} MLP on the train set. This choice aims to strengthen the model's ability to differentiate between classes. 
In CL, larger batch sizes often enhance performance by allowing more comparisons with negative examples, smoothing loss gradients. We found that a batch size of 256 suffices for good results. Extended training periods notably benefit CL. We train for 1000 epochs using a decoupled weight decay optimizer~\cite{adamw}. Parameter group learning rates are set with a cosine annealing schedule~\cite{cosine_annealing}. Our contrastive learned MLP (CL + MLP) consists of two linear layers: input size 768, sequentially transitioning to 768, and then to 128 with ReLU activation. After contrastive learning, the 32-dimensionality reduction MLP is applied.

\subsection{Ablation Study}
To assess the impact of incorporating CL, we compare the metrics of GAT with and without CL, alongside the MLP baseline. We train the MLP with CL to differentiate between answers only, leading to a new baseline MLP-A. This MLP is a two-layer model with hidden sizes 64 and 32, and ReLU activation. This comparison is excluded for the DeBERTa model, as MLP-A is trained solely on answers, and DeBERTa uses different embeddings, potentially leading to a distribution shift.

To further address our first research question, we evaluate the model's performance both with and without contrastive learning (CL). Table \ref{tab:add_cl} reveals significant improvements in GAT's performance with CL, particularly a remarkable 32\% increase in recall on the train set. While the validation set also shows overall improvements, there is a slight 3\% dip in precision, countered by an approximate 3\% increase in recall. Without CL, MLP's performance appears random. After using CL, there is an apparent improvement across all metrics. In particular, there is a 20\% improvement in both precision and recall for the train set. Validation recall sees an approximate 10\% increase, while precision increases by around 20\%.

\begin{table}[htbp]
\caption{Comparing performance between GAT, MLP, and kNN using contrastive learning (CL) versus without, with only answer (A) embeddings. For kNN we only show validation results. The best results for each metric and data split are highlighted in bold.}
\resizebox{\columnwidth}{!}{
\begin{tabular}{@{}lllll@{}}
\toprule
Split                  & Model                      & Recall   & Precision   & AUC-PR     \\ \midrule
\multirow{4}{*}{Train} & GAT                        & 0.5069 & 0.5844 & 0.4153  \\ 
                       & CL + GAT                   & \textbf{0.8244}   & \textbf{0.8281}   & \textbf{0.7118}      \\ 
                       & MLP-A                      & 0.2512 & 0.3123 & 0.2014  \\ 
                       & \multicolumn{1}{l}{CL + MLP-A} & 0.4286 & 0.5892 & 0.3987 \\ \midrule
\multirow{4}{*}{Val}   & GAT                        & 0.4972 & \textbf{0.5717} & 0.4096   \\ 
                       & CL + GAT                   & \textbf{0.5305}       & 0.5438          &  \textbf{0.4212}      \\ 
                       & MLP-A                      & 0.2256 & 0.3110 & 0.2057  \\ 
                       & \multicolumn{1}{l}{CL + MLP-A} & 0.3589 & 0.4956 & 0.3278  \\                        & \multicolumn{1}{l}{kNN}    & 0.2434       &  0.1895         &   0.2494      \\ \bottomrule
\end{tabular}
}
\label{tab:add_cl}
\end{table}

Despite CL significantly improving the results, the ablation study reveals it is not the only factor in improving performance. To answer our next research question \textit{2. Can we leverage graph structures to identify and learn these characteristics?}, we employ $k$-Nearest Neighbour (kNN) with CL-learned embeddings. We assess the independent expressiveness of these embeddings, anticipating that sufficiently robust embeddings would enable a reliable majority-voting mechanism. However, with $k=5$, kNN shows consistent underperformance (detailed in Table \ref{tab:add_cl}). Further exploration involved training the same MLP as introduced in Section \ref{sec:CL}, which showed improved performance compared to MLP without CL-learned embeddings. However, the MLP still trailed the performance of GAT, with approximately a 20\% decrease in validation recall, highlighting the significance of the graph structure. The attention mechanism of GAT is crucial in accurately identifying important neighbors. This refined approach which is solely reliant on spatial similarity outperformed the kNN method, highlighting the advantages of graph structures for efficient information propagation. Furthermore, edge masking ensures robustness by preventing information exchange between training and validation/test nodes during training. This method acts as a regularizer, enhancing the model's generalization capabilities~\cite{rong2020dropedge}.

% \begin{table}[h]
% \caption{Performance metrics comparison of GAT, MLP, and kNN all using contrastive learned embeddings. For kNN we only show validation results, as this method does not involve training. The best results for each metric and data split are highlighted in bold.}
% \begin{tabular}{@{}lllll@{}}
% \toprule
% Split                  & Model                      & Recall & Precision & AUC-PR  \\ \midrule
% \multirow{2}{*}{Train} & CL + GAT                   & \textbf{0.8244}   & \textbf{0.8281}   & \textbf{0.7118}       \\ 
%                        & CL + MLP-A                 & 0.4286 & 0.5892 & 0.3987     \\ \midrule
% \multirow{3}{*}{Val}   & CL + GAT                    & \textbf{0.5305}       & \textbf{0.5438}          &  \textbf{0.4212}         \\ 
%                        & CL + MLP-A                 & 0.3589 & 0.4956 & 0.3278     \\ 
%                        & \multicolumn{1}{l}{kNN}    & 0.2434       &  0.1895         &   0.2494      \\ \bottomrule
% \end{tabular}
% \label{tab:ablation}
% \end{table}

\subsection{Test Set Performance}
Finally, to address the research question \textit{3. If learned, can we use this knowledge to identify hallucinations among new incoming LLM generations through label recovery?}, we analyze the best-performing models by validation recall. The models that showcase the highest performance are GAT with and without contrastive learning. To ensure a fair comparison, we also consider the performance of the third-best model on the test set. The results are shown in Table \ref{tab:final}. Performance on the test set reveals that GAT with CL outperforms the other models on every metric except precision. The GAT structure proves crucial for higher recall.

\begin{table}[htbp]
    \caption{Comparing performance on the test set between the best performing models: GAT, MLP with CL, and GAT without CL. The best results for each metric and data split are highlighted in bold.}
    \resizebox{\columnwidth}{!}{
    \begin{tabular}{lccc}
        \toprule
         & Recall & Precision & AUC-PR \\
        \midrule
        CL + GAT &  \textbf{0.5142} & 0.5430 & \textbf{0.4057} \\
        GAT  & 0.4830 & \textbf{0.5603} & 0.3887 \\
        CL + MLP-A  & 0.3727  & 0.5122 &  0.3419 \\
        \bottomrule
    \end{tabular}
    }
    \label{tab:final}
\end{table}

\subsection{Generalizability on Other Benchmarks}
We assess the generalizability of our method on two real-world datasets, namely FEVER~\cite{Thorne18Fever} and SelfCheckGPT~\cite{selfcheckGPT}. Section \ref{sec:related_work} discusses their original applications, while Section \ref{ssec:benchmark} details how we modify the labels for our model.

%The FEVER dataset~\cite{Thorne18Fever} contains three types of claims, namely \textit{Supports}, \textit{Refutes}, and \textit{Not Enough Info}, each paired with evidence sentences. To apply our method to the FEVER dataset, we re-define the label as $y_i \in [0,1,2]$and generate the graph based on the train/val/test partitions of FEVER. The participating models~\cite{Thorne18Fever} formulate careful data processing approaches and make use of external sources to verify the factuality of the claims. If search-based evidence is found for a claim, it is classified as \textit{Supports} or \textit{Refutes}. Similarly, if no evidence is found, it is labeled as \textit{Not Enough Info}. % Moved to section 4.7
To benchmark our model's performance, we compare the results against the best performance of the first FEVER Shared Task challenge~\cite{thorne2018fact}, shown in Table \ref{tab:fever} as UNC-NLP. Our model outperforms UNC-NLP in precision, with accuracy being $4\%$ lower. However, it is important to stress that, in comparison, we solve a closed-book problem, avoiding the computational overload and necessity of any external data or search-based model.

\begin{table}[htbp]
\centering
\caption{For FEVER\cite{Thorne18Fever}: Performance metrics on the FEVER dataset. The best results are highlighted in bold.}
\label{tab:fever}
\resizebox{\columnwidth}{!}{
\begin{tabular}{@{}lccc@{}}
\toprule
Method   & \multicolumn{1}{l}{Recall} & \multicolumn{1}{l}{Precision} & \multicolumn{1}{l}{Label Accuracy} \\ \midrule
CL + GAT & 0.7079                     & \textbf{0.4712}                        & 0.6471                             \\
UNC-NLP & \textbf{0.7091}                     & 0.4227                        &  \textbf{0.6821}          \\ \bottomrule                    
\end{tabular}
}
\end{table}

% SelfCheckGPT~\cite{selfcheckGPT} contains 1908 sentences that are categorized as being \textit{Accurate},  \textit{Minor inaccurate}, and \textit{Major inaccurate}. 
Following the methodology in SelfCheckGPT~\cite{selfcheckGPT}, we use DeBERTa-large for sentence embeddings.
Our model falls short of the pairwise consistency metrics computed using DeBERTa-large embeddings (with BERTScore), as demonstrated in Table \ref{tab:selfcheckgpt}. A plausible explanation is the dataset's small size. The method with BERTScore needs multiple LLM-generated statements, while our method, trained on a small set, requires more examples for effective learning.

\begin{table}[htbp]
\caption{For SelfCheckGPT\cite{selfcheckGPT}: Factual sentences are labelled as \textit{Accurate}, NonFactual sentences are labelled as \textit{Major}- and \textit{Minor-inaccurate}. AUC-PR scores for Random and w/ BERTScore are computed on the entire dataset; our method's scores are calculated on the test set. The best results are highlighted in bold.}
\resizebox{\columnwidth}{!}{
\begin{tabular}{@{}lcc@{}}
\toprule
\multirow{2}{*}{Method} & \multicolumn{2}{c}{Sentence-level (AUC-PR)} \\
                                    & NonFactual              & Factual              \\ \midrule
Random          & 0.7296                & 0.2704                \\
LLM + BERT Scores    & \textbf{0.8196}               & \textbf{0.4423}                \\
CL + GAT                            & 0.7799                & 0.4002                \\ \bottomrule
\end{tabular}
}
\label{tab:selfcheckgpt}
\end{table}

%% file: content/conclusion.tex
This study shows the potential of GAT in LLM hallucination detection. Its adaptability and capabilities to find underlying graphical structures provide a significant advantage in discriminating between real and hallucinated generations. In the realm of hallucination detection, where information interconnects in complex ways, GATs' proficiency in navigating these connections proves invaluable. 
%Their ability to analyze interconnected data allows for a more comprehensive understanding of context, relationships, and potential influences, crucial in distinguishing between accurate and misleading content. 

Overall, this research reveals the pivotal role of structural information within graphs in discriminating between true and hallucinated statements. The incorporation of non-local aggregation serves to fortify these connections. The integration of a contrastive learned embedder enhances the discernment between true and hallucinated statements. Furthermore, this framework exhibits the capacity to extend beyond initial data, enabling generalization to real hallucinations. 

% Ideas for future work
\paragraph{Limitations and Future Work} Several limitations to our approach should be considered: 1) it requires effort to model the data, create ordinal categorical labels, and construct the graph structure; 2) it does not allow for transparency at all; 3) the method is difficult to scale, as adding nodes involves an exponential number of embedding comparisons for adding edges in the graph. Moreover, GATs face a limitation when adding new nodes to the graph, hindering real-time classification. Addressing these limitations could be a focus for future work, with the exploration of dynamic graph attention networks \cite{dynamicGAT} offering potential solutions. Dynamic GATs may facilitate the addition of new nodes, enabling real-time adaptation to evolving graph structures and addressing the current impracticality of real-time classification. Moreover, while the idea of this research was to model semantic relationships between individual utterances without explicitly assuming any connection between retrieval-based true statements and generations, it would be interesting to also simulate the query-answer baseline setting with the attention mechanism and analyze how the performance of our model changes.

% let's have something about scaling the graph; whenever you add new expriments, it would be nice to not re-create the graph each time, but to slowly build one big graph; because probably scaling it will boost performance -> you may already have something about this, but I cannot recall right now

% An important limitation to mention is that adding new nodes to the graph is not trivial in GATs since appending new nodes would require the network to recompute attention scores. Therefore, currently, it is not possible to use the model for real-time classification. For future work, it would be beneficial to look at dynamic graph attention networks \cite{dynamicGAT}. This would also enable a thorough investigation into data leakage between train, validation, and test sets. 
% Another limitation that needs to be addressed is the data used in this research. Since we intentionally prompt the LLM to generate deceptive statements, it might not be the same as genuine hallucinations that occur naturally. Therefore, the generalizability of our model on genuine hallucination data must be tested. For this, we propose to create new datasets of genuine hallucinations using already established LLM benchmarks such as TruthfulQA \cite{truth}.